%% file: main.tex
\documentclass[10pt,twocolumn,letterpaper]{article}

\usepackage[pagenumbers]{cvpr} 

\usepackage{graphicx}
\usepackage{amsmath, nccmath}
\usepackage{amssymb}
\usepackage{booktabs}

\def\eg{\emph{e.g}.,}
\def\ie{\emph{i.e}.}
\def\etal{\emph{et al.}}

\usepackage{amsmath}  
\usepackage{multirow}
\usepackage{tabularx}

\newcommand{\equref}[1]{(\ref{#1})}
\newcommand{\figref}[1]{Fig. \ref{#1}}
\newcommand{\tabref}[1]{Table \ref{#1}}

\newcommand\blfootnote[1]{%
  \begingroup
  \renewcommand\thefootnote{}\footnote{#1}%
  \addtocounter{footnote}{-1}%
  \endgroup
}

\usepackage[pagebackref,breaklinks,colorlinks]{hyperref}

\usepackage[capitalize]{cleveref}
\crefname{section}{Sec.}{Secs.}
\Crefname{section}{Section}{Sections}
\Crefname{table}{Table}{Tables}
\crefname{table}{Tab.}{Tabs.}

\def\cvprPaperID{10469} 
\def\confName{CVPR}
\def\confYear{2023}

\begin{document}

\title{PartMix: Regularization Strategy to Learn Part Discovery for Visible-Infrared \\ Person Re-identification}

\author{
Minsu Kim\textsuperscript{\rm 1},
Seungryong Kim\textsuperscript{\rm 2},
Jungin Park\textsuperscript{\rm 1},
Seongheon Park\textsuperscript{\rm 1},
Kwanghoon Sohn\textsuperscript{\rm 1,3}\thanks{Corresponding author} \\
\textsuperscript{\rm 1} Yonsei University
\textsuperscript{\rm 2} Korea University
\textsuperscript{\rm 3} Korea Institute of Science and Technology (KIST)\\
{\tt\small \{minsukim320, newrun, sam121796, khsohn\}@yonsei.ac.kr seungryong\_kim@korea.ac.kr 
}
}
\maketitle
\blfootnote{This research was supported by the National Research Foundation of Korea (NRF) grant funded by the Korea government (MSIP) (NRF2021R1A2C2006703).}

\begin{abstract}
Modern data augmentation using a mixture-based technique can regularize the models from overfitting to the training data in various computer vision applications, but a proper data augmentation technique tailored for the part-based Visible-Infrared person Re-IDentification (VI-ReID) models remains unexplored.
In this paper, we present a novel data augmentation technique, dubbed \textbf{PartMix}, that synthesizes the augmented samples by mixing the part descriptors across the modalities to improve the performance of part-based VI-ReID models.
Especially, we synthesize the positive and negative samples within the same and across different identities and regularize the backbone model through contrastive learning. 
In addition, we also present an entropy-based mining strategy to weaken the adverse impact of unreliable positive and negative samples.
When incorporated into existing part-based VI-ReID model, PartMix consistently boosts the performance. 
We conduct experiments to demonstrate the effectiveness of our PartMix over the existing VI-ReID methods and provide ablation studies.
\end{abstract}
\vspace{-8pt}


\section{Introduction}
\label{sec:intro}
Person Re-IDentification (ReID), aiming to match person images in a query set to ones in a gallery set captured by non-overlapping cameras, has recently received substantial attention in numerous computer vision applications, including video surveillance, security, and persons analysis~\cite{Ye21_2, Zheng16}.
Many ReID approaches~\cite{Liao15, Matsukawa16, Chen18, Sun18, Li18, Yang19, Zhang19, Chen19, Zhu20} formulate the task as a visible-modality retrieval problem, which may fail to achieve satisfactory results under poor illumination conditions. 
To address this, most surveillance systems use an infrared camera that can capture the scene even in low-light conditions. However, directly matching these infrared images to visible ones for ReID poses additional challenges due to an inter-modality variation~\cite{Ye19, Ye20, Ye20_2}.

 \begin{figure}[t]
        \centering
            \begin{subfigure}{1\linewidth}
            {\includegraphics[width=1\linewidth]{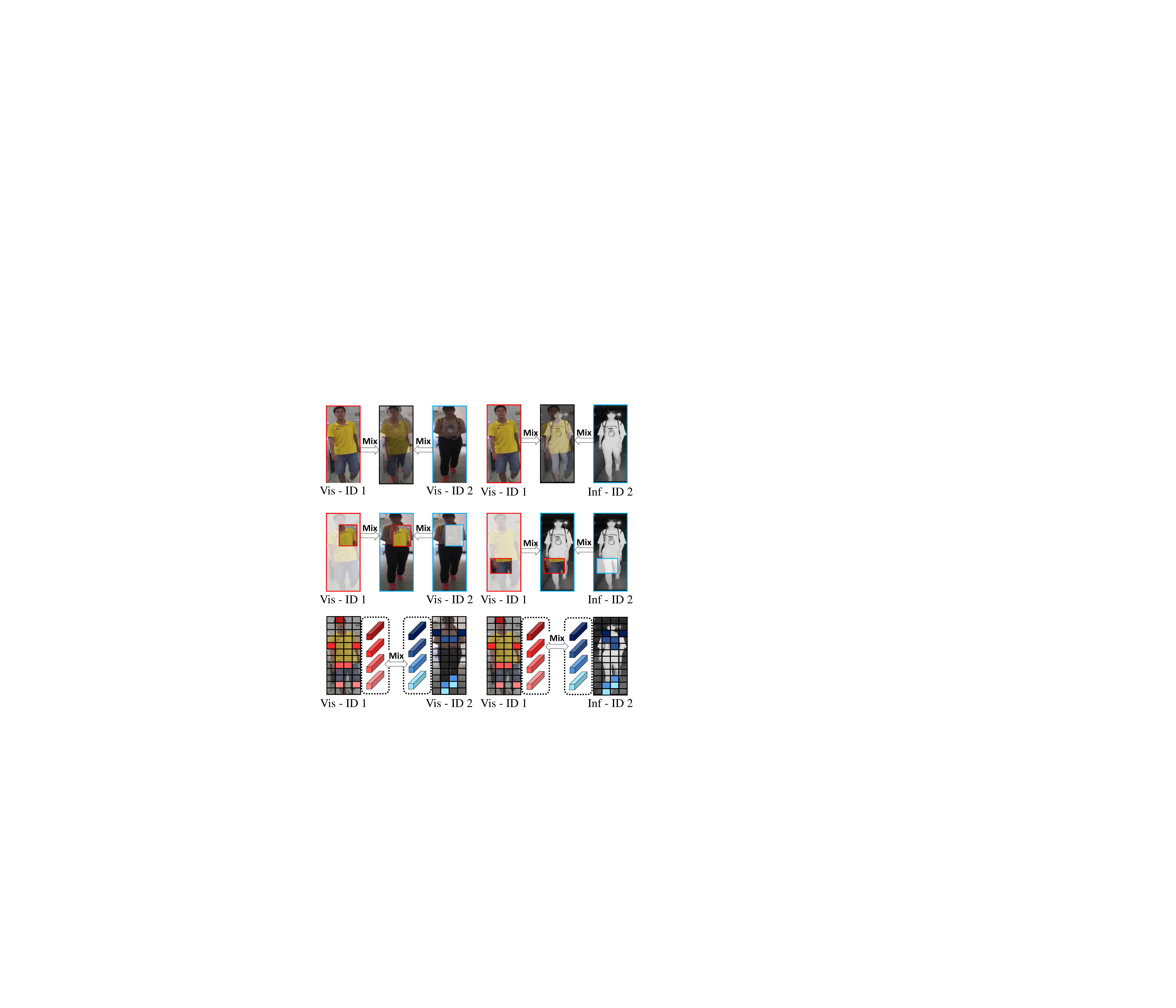}}
            \caption{MixUp~\cite{Zhang18}}
            \label{fig:2a}
           \end{subfigure}\\
           \vspace{+5pt}
           \begin{subfigure}{1\linewidth}
            {\includegraphics[width=1\linewidth]{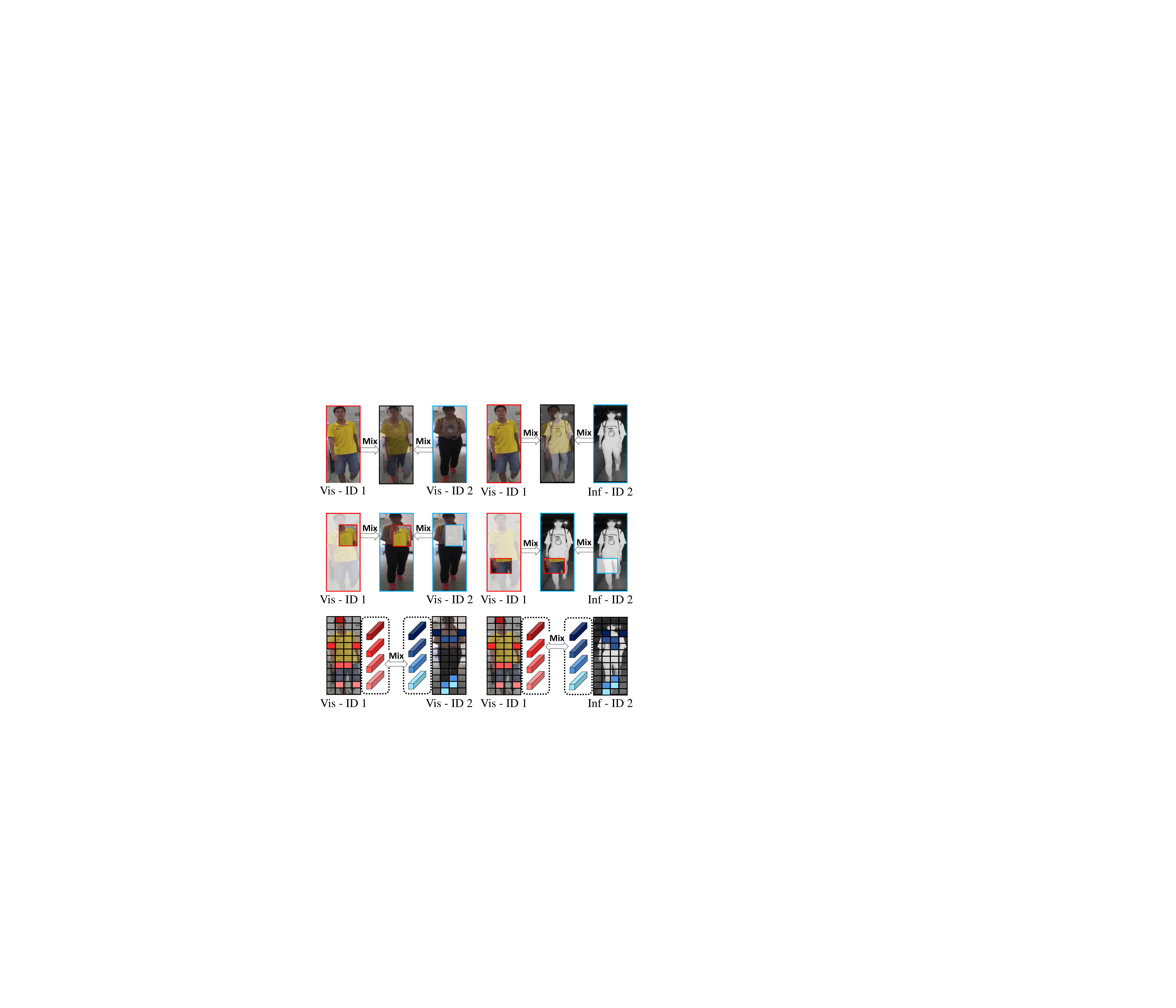}}
            \caption{{CutMix}~\cite{Yun19}}
            \label{fig:2b}
           \end{subfigure}  \\
           \vspace{+5pt}
           \begin{subfigure}{1\linewidth}
            {\includegraphics[width=1\linewidth]{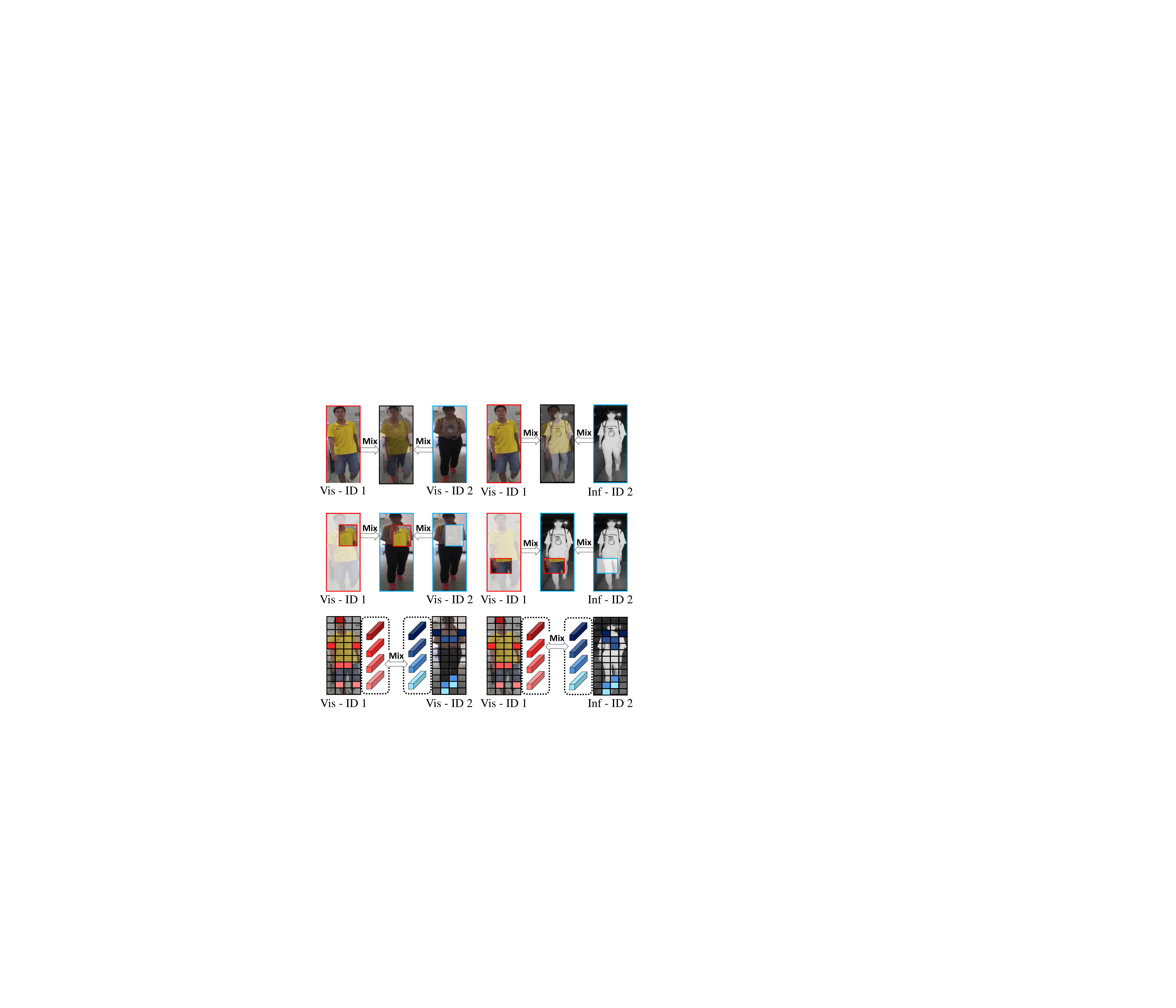}}
            \caption{PartMix (Ours)}
            \label{fig:2c}
           \end{subfigure} \\ 
           \vspace{-5pt}
        \caption{\textbf{Comparison of data augmentation methods for VI-ReID.}
        (a) MixUp~\cite{Zhang18} using a global image mixture and (b) CutMix~\cite{Yun19} using a local image mixture can be used to regularize a model for VI-ReID, but these methods provide limited performances because they yield unnatural patterns or local patches with only background or single human part. Unlike them, we present (c) PartMix using a part descriptor mixing strategy, which boosts the VI-ReID performance (Best viewed in color).
        }
    \label{fig:f3}
    \vspace{-9pt}
    \end{figure}
    
To alleviate these inherent challenges, Visible-Infrared person Re-IDentification (VI-ReID)~\cite{Dai18, Li20, Wei21, Wang19_2, Fan20, Ye18, Feng19, Ye19, Ye20_2, Ye18_2, Hao19, Park21} has been popularly proposed to handle the large intra- and inter-modality variations between visible images and their infrared counterparts.
Formally, these approaches first extract a person representation from whole visible and infrared images, respectively, and then learn a modality-invariant feature representation using feature alignment techniques, \eg{} triplet~\cite{Ye18, Dai18, Feng19, Ye19, Li20, Ye20_2} or ranking criterion~\cite{Ye18_2, Hao19}, so as to remove the inter-modality variation.
However, these global feature representations solely focus on the most discriminative part while ignoring the diverse parts which are helpful to distinguish the person identity~\cite{Ye20_3, Wei21_2}.

Recent approaches~\cite{Ye20_3,Wei21_2, Wu21} attempted to further enhance the discriminative power of person representation for VI-ReID by capturing diverse human body parts across different modalities.
Typically, they first capture several human parts through, e.g., horizontal stripes~\cite{Ye20_3}, clustering~\cite{Wei21_2}, or attention mechanisms~\cite{Wu21} from both visible and infrared images, extract the features from these human parts, and then reduce inter-modality variation in a part-level feature representation. Although these methods reduce inter-modality variation through the final prediction (e.g., identity probability), learning such part detector still leads to overfitting to the specific part because the model mainly focuses on the most discriminative part to classify the identity, as demonstrated in~\cite{Hou18, Wei17, Choe19, Mai20}. 
In addition, these parts are different depending on the modality, it accumulates errors in the subsequent inter-modality alignment process, which hinders the generalization ability on unseen identity in test set.

On the other hand, many data augmentation~\cite{Zhang18, Yun19, Walawalkar20, Kim20_2, Lee20, Shen22} enlarge the training set through the image mixture technique~\cite{Zhang18}.
They typically exploit the samples that linearly interpolate the global~\cite{Zhang18, Verma19, Shen22} or local~\cite{Yun19, Kim20_2} images and label pairs for training, allowing the model to have smoother decision boundaries that reduce overfitting to the training samples.
This framework also can be a promising solution to reduce inter-modality variation by mixing the different modality samples to mitigate overfitting to the specific modality, but directly applying these techniques to part-based VI-ReID models is challenging in that they inherit the limitation of global and local image mixture methods (\eg{} ambiguous and unnatural patterns, and local patches with only background or single human part).
Therefore, the performance of part-based VI-ReID with these existing augmentations would be degraded.

In this paper, we propose a novel data augmentation technique for VI-ReID task, called \textbf{PartMix}, that synthesizes the part-aware augmented samples by mixing the part descriptors.
Based on the observation that learning with the unseen combination of human parts may help better regularize the VI-ReID model, we randomly mix the inter- and intra-modality part descriptors 
to generate positive and negative samples within the same and across different identities, and regularize the model through the contrastive learning.
In addition, we also present an entropy-based mining strategy to weaken the adverse impact of unreliable positive and negative samples. 
We demonstrate the effectiveness of our method on several benchmarks~\cite{Wu17, Nguyen17}.
We also provide an extensive ablation study to validate and analyze components in our model.


\section{Related work}
\label{sec:related}
\paragraph{Person ReID.}
Person Re-IDentification (ReID) aims to search a target person from a large gallery set, where the images are captured from non-overlapping visible camera views.
With the advent of deep convolutional neural networks (CNNs), 
to solve this task, existing works~\cite{Wang16, Chen17, Chen18} encourage the person representation within the same identity to be similar through feature-level constraint, including triplet constraint~\cite{Wang16}, quadruplet constraint~\cite{Chen17}, or group consistency constraint~\cite{Chen18}.
However, since these methods learn features from the whole person image, they often suffer from intra-modality variation caused by the human pose variation and part occlusions~\cite{Sun18}.
Thus, there have been many efforts to focus on extracting human body parts that can provide fine-grained person image descriptions through uniform partitions~\cite{Varior16, Sun18, Fu19, Wang18, Zheng19} or attention mechanism~\cite{Li18, Zhao17, Liu17, Liu17_2, Yang19, Zheng19_2}.

\begin{figure*}
   \centering
   \includegraphics[width=1 \linewidth]{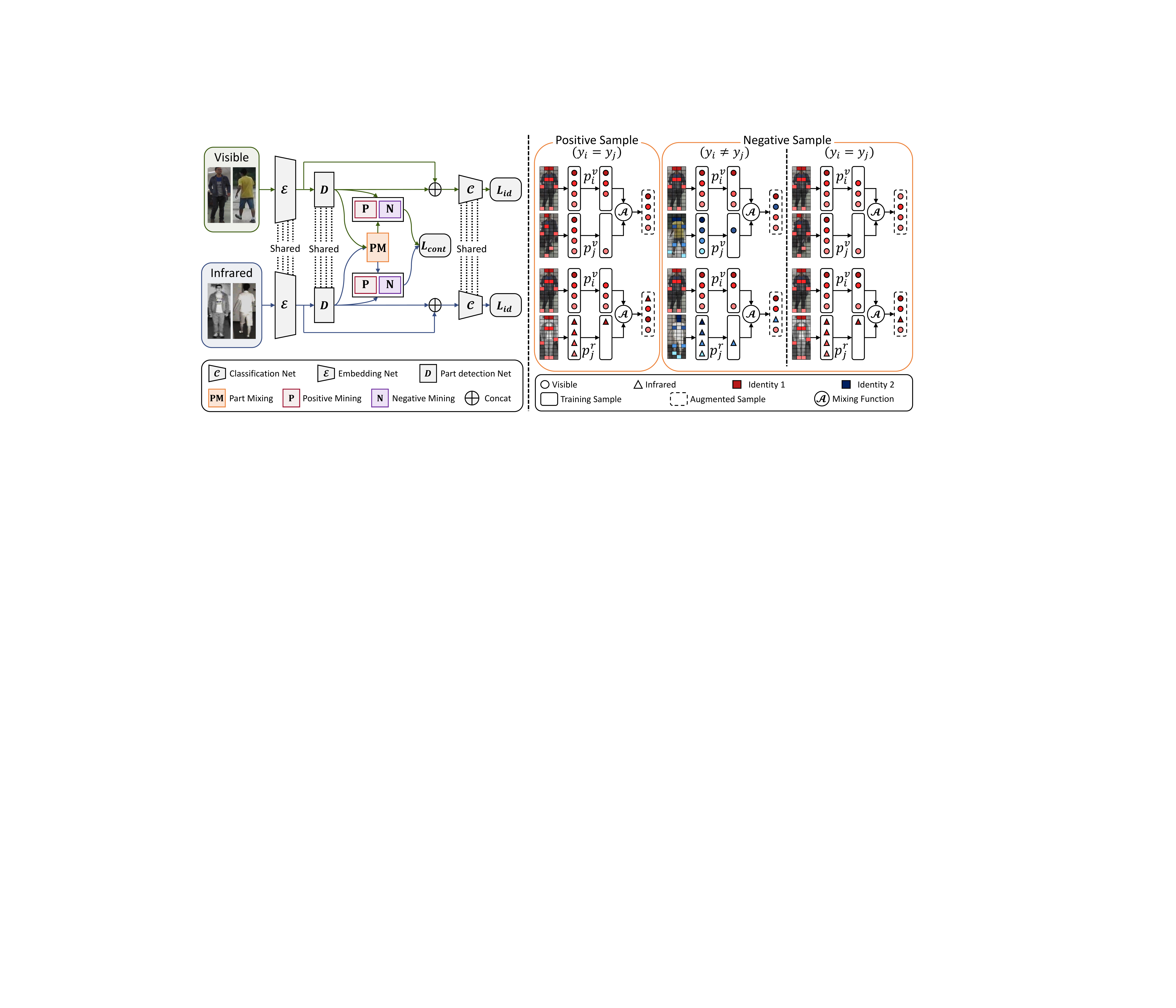}
   \vspace{-10pt}
   \caption{\textbf{Overview of our PartMix-based framework for VI-ReID.} The visible and infrared images are fed into the embedding network to extract features and obtain part descriptors through the global average pooling layer.
The part descriptors from visible and infrared modalities are fed into the part mixing module to synthesize the positive and negative samples.
The reliable positive and negative samples are selected through the positive and negative mining module for contrastive learning.}
\label{fig:f2_1}
\vspace{-9pt}
\end{figure*}

\vspace{-9pt}
\paragraph{Visible-Infrared Person ReID.}
Most surveillance systems deploy infrared images to achieve satisfactory results under poor illumination conditions~\cite{Kim19, Wu17}.
However, directly applying person re-id methods suffer from the different distribution between modalities, typically known as modality discrepancy~\cite{Ye19, Ye20, Ye20_2}.
To alleviate this, Visible-Infrared person re-id (VI-ReID) has been popularly studied to match a person between visible and infrared images with challenges posed by large intra-modality variation and inter-modality variations. 
Wu \etal{}~\cite{Wu17} first introduced a large-scale VI-ReID dataset, named SYSU-MM01, and proposed a deep zero-padding strategy to explore modality-specific structure in a one-stream network. 
Recently, modality invariant feature learning based VI-ReID has been proposed to project the features from different modalities into the same feature space.
Formally, these methods extract features from visible and infrared person images and then reduce the inter-modality discrepancy by feature-level constraints, such as triplet constraint~\cite{Ye18, Feng19, Ye19, Ye20_2}, or ranking constraint~\cite{Ye18_2, Hao19}.
However, these methods usually focus only on the most discriminative part rather than the diverse parts which are helpful to distinguish different persons~\cite{Ye20_3, Wei21_2}.
Therefore, several methods have been proposed to align inter-modality discrepancy in a fine-grained manner, exploiting horizontal stripes~\cite{Ye20_3}, or inter-modality nuances~\cite{Wu21, Wei21_2}.
However, the part detector can be easily overfitted to the specific part.
Moreover, these part maps are not semantically consistent across modalities, showing limited performance.

\vspace{-9pt}
\paragraph{Data Augmentation.}
Data augmentation has been widely explored in various computer vision tasks~\cite{Shorten19, Xie20, Zhang18, Hendrycks19, Verma19, Lee21} to prevent deep neural network from overfitting to the training data.
Specifically, mixture-based methods, including global image mixture~\cite{Zhang18, Verma19, Shen22}, e.g., MixUp, and local image mixture~\cite{Yun19, Kim20_2}, e.g., CutMix, are dominant regularization strategies. 
These methods formally synthesize the virtual training samples by linearly interpolating the images and the corresponding labels. 
It results in smoother decision boundaries that reduce overfitting to the training data.
On the other hand, some methods~\cite{Lee20, Shen22} leveraged the different image mixtures for contrastive-based unsupervised learning.
Recently, there exist several efforts to adapt the global image mixture to VI-ReID tasks~\cite{Zhang21_2, Huang22}, but there still remains the problem of unnatural patterns, which hinder the localization ability~\cite{Yun19, Venkataramanan22} for part-based VI-ReID.
Unlike the methods above, we present for the first time a part-level augmentation for part-based VI-ReID, overcoming the limitations of existing data augmentation methods~\cite{Zhang18, Verma19, Shen22, Yun19}.


\section{Proposed Method}
\label{sec:proposed}
\subsection{Preliminaries and Problem Formulation}
Let us denote visible and infrared image sets as $\mathcal{X}^{v} =\{x^{v}_{i}\}_{i=1}^{N^{v}}$ and $\mathcal{X}^{r}=\{x^{r}_{i}\}_{i=1}^{N^{r}}$, where $x^{v}_{i}$ and $x^{r}_{i}$ are images, and $N^{v}$ and $N^{r}$ are the number of visible and infrared images, and they are unpaired.
For visible and infrared image sets, the corresponding identity label sets are defined such that $\mathcal{Y}^{v}=\{y^{v}_{i}\}^{N^{v}}_{i = 1}$ and $\mathcal{Y}^{r}=\{y^{r}_{i}\}^{N^{r}}_{i = 1}$, whose label candidates are shared.
The objective of VI-ReID is to learn modality-invariant person descriptors, denoted by $d^{v}$ and $d^{r}$, for matching persons observed from visible and infrared cameras. For simplicity, we denote visible and infrared modalities as $t \in \{v, r\}$ unless stated.

Most recent state-of-the-art methods are based on part-based person representation~\cite{Ye20_3, Wei21_2, Wu21} that aims to extract discriminative human parts information and use them to enhance the robustness of person representation against human pose variation across the modality.
These methods typically involve discovering diverse and discriminative human parts in an attention mechanism, and generating the person descriptor by assembling a global descriptor and part descriptors for retrieving visible (or infrared) images according to the given infrared (or visible) images. 

Specifically, given visible and infrared images, the feature map for each modality is computed through an embedding network $\mathcal{E}(\cdot)$ such that $f^{t} = \mathcal{E}(x^{t})$. 
The part detector $\mathcal{D}(\cdot)$ then produces human parts, followed by sigmoid function $\sigma(\cdot)$, to output 
part map probability, denoted by 
$\{m^{t}({k})\}^{M}_{k=1} = \sigma(\mathcal{D}(f^{t}))$, where $M$ is the number of part maps.
The part descriptors are then formulated as follows: 
\begin{equation}\label{equ:1}
    p^{t}=[p^{t}(k)]_{k=1}^{M}=[\mathrm{GAP}(m^{t}(k) \odot f^{t})]_{k=1}^{M},
\end{equation}
where $\mathrm{GAP}(\cdot)$ denotes a global average pooling, $\odot$ is an element-wise multiplication, and $[\cdot]$ is a concatenate operation. Note that they apply element-wise multiplication between $m^{t}(k)$ and each channel dimension in $f^{t}$.
They finally concatenate the global descriptor $g^{t}$ such that $l^{t} = \mathrm{GAP}(f^{t})$ and part descriptors $p^{t}$ to obtain person descriptor $d^{t}$ for matching the persons observed from visible and infrared cameras such that
\begin{equation}\label{equ:2}
     d^{t} = [g^{t}, p^{t}].
\end{equation}

To train such a model, since only identity labels are available, they adopted a cross-entropy loss between the identity probabilities and ground-truth identities.
In addition, they also adopted several loss functions, including knowledge distillation~\cite{Wu21} or metric learning~\cite{Ye20_3, Wu21, Wei21_2} loss, to learn modality invariance in a part-level feature representation.

\begin{figure*}[t]
	\centering
	\includegraphics[width=0.95 \linewidth]{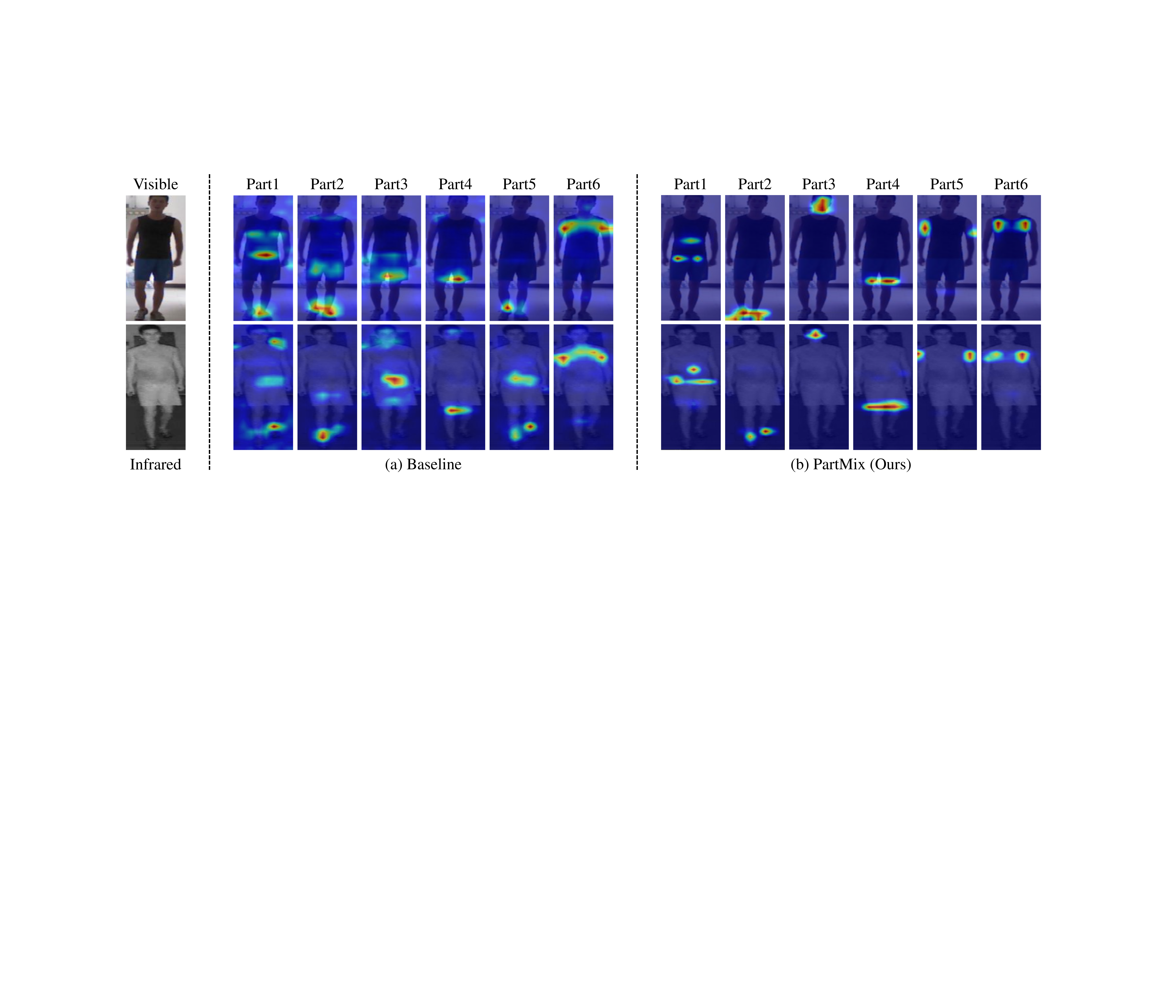}
        \vspace{-3pt}	
 \caption{\textbf{Visualization of part detection results by (a) baseline~\cite{Wu21} and (b) baseline with PartMix (Ours).} 
	The baseline method is easily overfitted to the specific part, while our method can capture diverse and discriminative human parts. 
	}
	\label{fig:f4}
	\vspace{-9pt}
\end{figure*}

While these losses let the network focus on human parts across modalities and reduce inter-modality variations by aligning person descriptors within the same identity, the part detectors learned by these methods have been often overfitted to the specific part, as exemplified in~\figref{fig:f4}.
In addition, these learned parts may not be semantically consistent across modalities~\cite{Hou21, Inoue18, Li16}.
Therefore, they fail to localize discriminative and semantically-consistent human parts across the visual and infrared modalities, thus showing limited performance.
\begin{figure}
   \centering
   \includegraphics[width=1 \linewidth]{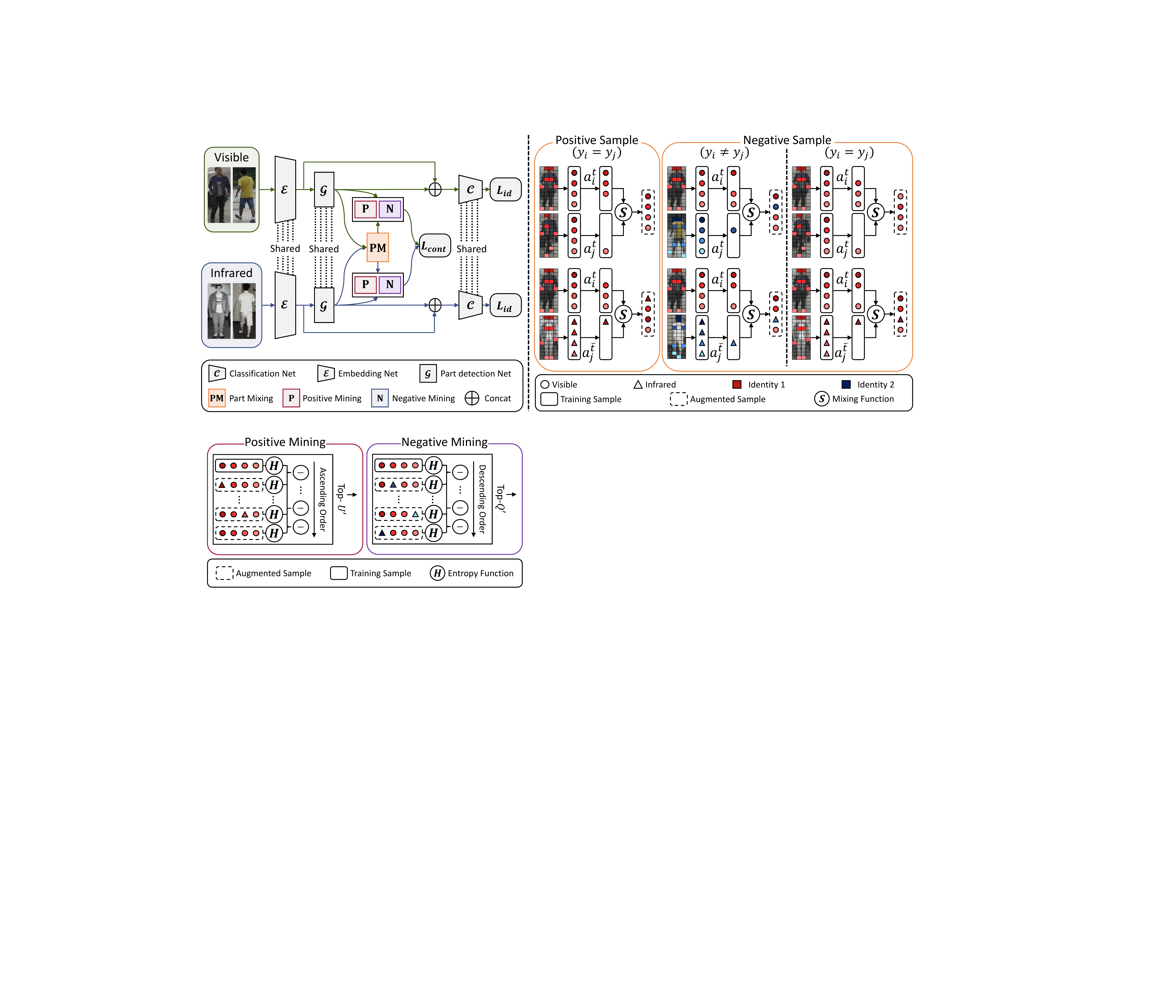}
   \vspace{-10pt}
   \caption{\textbf{Illustration of entropy-based mining strategy.} 
   It computes the difference between the pairwise entropy for positive and negative samples. These uncertainty value sets are sorted in ascending and descending order to select reliable positive and negative samples.}
\label{fig:f2_2}
\vspace{-9pt}
\end{figure}

\subsection{Overview}
To overcome the limitations of previous part-based person representation methods~\cite{Ye20_3, Wei21_2, Wu21}, our PartMix, which we introduce in the remainder of this section, accounts for the observation that learning with the part-aware augmented samples across both inter- and intra-modality can help better regularize the model. Unlike conventional methods~\cite{Zhang18, Verma19, Shen22, Yun19, Kim20_2} that exploit image mixture techniques that yield unnatural patterns, and local patches with only background or  single human part,
we present a novel part mixing strategy to synthesize augmented samples by mixing partial descriptors across the modalities, and use them to synthesize positive and negative samples to maximize the similarities of positive pairs and minimize the similarities of negative pairs through the contrastive learning objective.
It helps to regularize the model and mitigate the overfitting to the specific part and modality, improving the generalization capability of the model.
Furthermore, to eliminate the unreliable positive and negative samples, we present an entropy-based mining strategy that can help guide a representation to be more discriminative.

\subsection{Part Mixing for Data Augmentation}
One of the most straightforward ways of leveraging regularization to better learn part discovery may be to utilize existing augmentation techniques, e.g., using global~\cite{Zhang18, Verma19, Shen22} or local~\cite{Yun19, Kim20_2} image mixtures. 
These existing strategies, however, are difficult to be directly applied to part-based VI-ReID methods~\cite{Ye20_3, Wei21_2, Wu21} due to the following two respects. Firstly, directly applying the global image mixture methods~\cite{Zhang18, Verma19, Shen22} suffers from the locally ambiguous and unnatural pattern, and thus mixed sample confuses the model, especially for localization as demonstrated in~\cite{Yun19, Venkataramanan22}.
Secondly, mixing the local image region~\cite{Yun19, Kim20_2} without part annotation may contain only a background or single human part, and thus it may cause performance degradation for part-based models which require diverse and discriminative discovery of human parts for distinguishing the person identity.

To overcome these, we present a novel data augmentation technique tailored to part-based methods, called PartMix, that mixes the part descriptors extracted from different person images.
By mixing the part descriptors rather than the images, we can synthesize the augmented samples with diverse combinations of human parts. 
Concretely, we first collect the part descriptors in the visible and infrared modalities of the mini-batch, denoted as the descriptor bank $P^{t} = \{ p^{t}_{1}, p^{t}_{2} ..., p^{t}_{N^{t}}\}$.
We then mix the part descriptors through part mix operation across the inter-modality $
\mathcal{A}(p^{v}_{i}(u), p^{r}_{j}(h))$ and intra-modality $\mathcal{A}(p^{t}_{i}(u), p^{t}_{j}(h))$ sequentially as follows: 
\begin{equation}\label{equ:4}
\begin{split}
&\mathcal{A}(p^{v}_{i}(u), p^{r}_{j}(h)) \\ 
&\;\;\;\;\; =  [p^{v}_{i}(1),...,p^{v}_{i}(u-1),p^{r}_{j}(h), p^{v}_{i}(u+1),...,p^{v}_{i}(M)], \\
&\mathcal{A}(p^{t}_{i}(u), p^{t}_{j}(h)) \\ 
&\;\;\;\;\; =  [p^{t}_{i}(1),...,p^{t}_{i}(u-1),p^{t}_{j}(h), p^{t}_{i}(u+1),..., p^{t}_{i}(M)], 
\end{split}
\end{equation}
where 
$h, u$ denote the randomly sampled indexes of part descriptors $p^{t}$.
Note that we exclude the global descriptor $g^{t}$ in the part mixing above because it contains all human body parts information. 

\subsection{Sample Generation for Contrastive Learning}
Existing image mixture-based methods~\cite{Zhang18, Verma19, Shen22, Yun19, Kim20_2} generate the training samples by linearly interpolating the images and the corresponding labels. 
These approaches, however, only synthesize the samples with the combination of identities in the training set, and thus they have limited generalization ability on the VI-ReID task where identities in the testing set are different from the training set. 
To alleviate this, we present a sample generation strategy that can synthesize positive and negative samples with the unseen combination of human parts (\ie{} the unseen identity).
In the following section, we explain how to achieve positive bank $B^{+,t}_{i}$ and negative bank $B^{-,t}_{i}$ in detail.
For simplicity, only visible samples are described as an example.

\vspace{-10pt}
\paragraph{Positive Samples.}
Our first insight is that the combination of the human parts of the persons with the same identity has to be consistent.
To this end, we design positive samples that mix the same part information between the person images within the same identity.
Specifically, we mix the part descriptors with the same identity using \eqref{equ:4}.
Each positive sample for visible modality is denoted as 
\begin{equation}\label{equ:5}
b^{+,v}_{i} = [\mathcal{A}(p^{v}_{i}(k), p^{r}_{j}(k)), \mathcal{A}(p^{v}_{i}(k), p^{v}_{j}(k))],
\\
\;\;\text{if} \;y_{i}=y_{j}.
\end{equation}
Note that we only mix the part-level descriptor within the same identity (\ie{} $y_{i}=y_{j}$). 

\vspace{-10pt}
\paragraph{Negative Sample.}
The positive samples in \eqref{equ:5} encourage the part-level feature within the same identity across the inter- and intra-modality to be invariant. 
This, however, does not guarantee that the model can distinguish the person identity with different combinations of human parts, and localize the diverse human parts within the same person.

To overcome this, we design the negative samples that encourage the model to distinguish the person identity when the combination of human parts is different and localize diverse human parts in each identity.
We mix the part descriptor within the same and different identity using \equref{equ:4} as
\begin{equation}\label{equ:6}
b^{-,v}_{i} = 
\left\{\begin{matrix}
[\mathcal{A}(p^{v}_{i}(k), p^{r}_{j}(h)), \mathcal{A}(p^{v}_{i}(k), p^{v}_{j}(h))], \; \text{if} \;y_{i} = y_{j}\\
[\mathcal{A}(p^{v}_{i}(k), p^{r}_{j}(k)), \mathcal{A}(p^{v}_{i}(k), p^{v}_{j}(k))], \; \text{if} \;y_{i} \neq y_{j}
\end{matrix}\right.
, \end{equation}
where $k$ and $h$ denote the different indexes of part descriptors.
Note that our negative samples cover the unseen combination of human parts in the training set. 
Therefore, these samples can be seen as out-of-distribution negative samples that can provide supportive information for improving the generalization capability of the model as investigated in~\cite{Sinha21, Geiping22}.

\subsection{Entropy-based Mining Strategy}
Even though the proposed sample generation strategy through part mixing yields comparable performance to some extent (which will be discussed in experiments), it relies on the assumption that there are different human part information (e.g. clothes, hairstyle) for each person.
However, several persons with different identities share a similar appearance (\eg{} wearing similar clothes). 
Therefore, simply mixing these samples makes false negative samples that still have the same combination of human parts. Moreover, the false positive samples can be synthesized by randomly mixing human parts that have different semantic meanings, especially in the early training stage.

To overcome this, inspired by the uncertainty-based approaches~\cite{Grandvalet04, Shannon48}, we present an entropy-based mining strategy for eliminating the false positive and false negative samples.
We exploit the difference between the pairwise entropy of the identity prediction that can be an uncertainty measure for positive and negative samples, where the reliable samples are determined as a positive pair with a smaller entropy difference and a negative pair with a larger entropy difference.
We first obtain identity probability for each sample through the part-level identity classifier $\mathcal{C}_{p}(\cdot)$.
We then compute the difference between the pairwise entropy for positive and negative samples as follows:

\begin{equation}\label{equ:7}
\begin{matrix}
 h^{+,v}_{i}= [ |H(\mathcal{C}_{p}(p^{v}_{i})) - H(\mathcal{C}_{p}(b^{+,v}_{i}(j)))| ]_{j = 1}^{U},
 \\ h^{-,v}_{i}= [ |H(\mathcal{C}_{p}(p^{v}_{i})) - H(\mathcal{C}_{p}(b^{-,v}_{i}(j)))| ]_{j = 1}^{Q},
\end{matrix}
\end{equation}
where $H(\cdot)$ is entropy function~\cite{Shannon48},
 and $U$, $Q$ are the number of positive and negative pairs.
 These uncertainty value sets for positive $h^{+,v}_{i}$
 and negative $h^{-,v}_{i}$ are then sorted in ascending and descending order, respectively.
We select the top ${U}'$ and ${Q}'$ samples for positive bank $B^{+,v}_{i} = [b^{+,v}_{i}(j)]^{{U}'}_{j=1}$ and negative bank $B^{-,v}_{i} = [b^{-,v}_{i}(j)]^{{Q}'}_{j=1}$, respectively.

\subsection{Loss Functions}\label{sec:36}
In this section, we describe several loss functions to train our network.
We newly present contrastive regularization loss $\mathcal{L}_\mathrm{cont}$ and part ID loss $\mathcal{L}_\mathrm{aid}$ to regularize the model through
positive and negative samples.

\vspace{-9pt}
\paragraph{Contrastive Regularization Loss.}
We aim to maximize the similarities between positive pairs and minimize similarities between negative pairs.
To this end, inspired by ~\cite{Wu21_2, Oord18}, we adopt the contrastive regularization loss that jointly leverages positive and negative samples to regularize the model, and thus the model can overcome the limitation of strong reliance on supervision signal.
We first define the similarity between part descriptors and positive (negative) samples, defined as $s^{+}_{i,j} = \mathrm{sim}(p_{i}, b^{+}_{i}(j))$
($s^{-}_{i,k} = \mathrm{sim}(p_{i}, b^{-}_{i}(k))$), where $\mathrm{sim}(\cdot, \cdot)$ denotes a similarity function.
The contrastive loss can be written as follows:
\begin{equation}\label{equ:8}
\begin{split}
  &\mathcal{L}_\mathrm{cont} = \\
  &\sum^{N^{v}+N^{r}}_{i=1}-\log \frac{\sum^{{U}'}_{j=1}\mathrm{exp}(s^{+}_{i,j} / \tau)}{\sum^{{U}'}_{j=1}\mathrm{exp}(s^{+}_{i,j} / \tau) + \sum^{{Q}'}_{k=1}\mathrm{exp}(s^{-}_{i,k} / \tau)}
  \end{split},
\end{equation}
where $\tau$ is a scaling temperature parameter.

\vspace{-9pt}
\paragraph{Part ID Loss.}
To estimate accurate entropy values for the augmented samples, we adopt part ID loss for an additional part classifier that provides identity scores from part descriptors.
It allows the model to eliminate the unreliable positive and negative samples, and learn discriminative and semantically consistent part discovery, simultaneously.
\begin{equation}\label{equ:9}
\begin{split}
\mathcal{L}_\mathrm{aid} = -\frac{1}{N} \sum^{N}_{i=1} y_{i} \log(\mathcal{C}_{p}(p_{i})).
\end{split}
\end{equation}

\vspace{-8pt}
\paragraph{Total Loss.}
Following the baseline~\cite{Wu21}, we also adopt 
modality learning loss $\mathcal{L}_{\mathrm{ML}}$, 
modality specific ID loss $\mathcal{L}_{\mathrm{sid}}$, center cluster loss $\mathcal{L}_{\mathrm{cc}}$, 
and identity classification loss $\mathcal{L}_{\mathrm{id}}$.
The detailed losses are described in the supplementary material.
The total loss function of our approach can be written as $\mathcal{L} = \mathcal{L}_{\mathrm{id}} + \mathcal{L}_{\mathrm{cc}}+ \lambda_{\mathrm{sid}}\mathcal{L}_{\mathrm{sid}} +\lambda_{\mathrm{ML}}\mathcal{L}_{\mathrm{ML}}+ \lambda_{\mathrm{aid}}\mathcal{L}_{\mathrm{aid}} + \lambda_{\mathrm{cont}}{L}_{\mathrm{cont}}$, where 
$\lambda_{\mathrm{sid}}$,
$\lambda_{\mathrm{ML}}$,  $\lambda_{\mathrm{aid}}$, and $\lambda_{\mathrm{cont}}$ are weights that control the importance of each loss.

\subsection{Discussion}
Most recent trends in inter-domain scenarios (\eg{} domain adaptation~\cite{Yue21_2, Zhang21}) exploited counterfactual intervention to learn domain invariant knowledge, improving generalization capabilities. 
These frameworks consist of two main components: generating ``counterfactual samples" by changing the domain (\eg{} style) and using them in the model training for ``intervention". 
Our PartMix satisfies these conditions, as we synthesize the modality-mixed samples by changing the part descriptor across the modality and training the whole model with these samples. 
Therefore, we can interpret our PartMix as a counterfactual intervention for inter-modality part-discovery, where the part mixing module can be viewed as a ``counterfactual” sample generator and the contrastive regularization as an ``intervention”. 
This allows our model to learn modality invariant part representation by encouraging the part discovery to be invariant under different interventions.


\section{Experiments}
\label{sec:exper}

\subsection{Experimental Setup}
In this section, we comprehensively analyze and evaluate our PartMix on several benchmarks~\cite{Wu17, Nguyen17}.
First, we analyze the effectiveness of our PartMix and comparison with other regularization methods.
We then evaluate our method compared to the state-of-the-art methods for VI-ReID. 
In the experiment, we utilize MPANet~\cite{Wu21} as our baseline model.
Additional implementation details will be explained in the supplementary material.

\vspace{-9pt}
\paragraph{Dataset.}
We evaluate our method on two benchmarks, SYSU-MM01~\cite{Wu17} and RegDB~\cite{Nguyen17}. Firstly, SYSU-MM01 dataset~\cite{Wu17} is a large-scale VI-ReID dataset.
This dataset contains 395 identities with 22,258 visible images acquired by four cameras and 11,909 near-infrared images acquired by two cameras for the training set.
The testing set contains 96 identities with 3,803 near-infrared images in the query, and 301 and 3,010 visible images in the gallery for single-shot and multi-shot, respectively. 
Secondly, RegDB dataset~\cite{Nguyen17} contains 4,120 visible and infrared paired images with 412 person identities, where each person has 10 visible and 10 far-infrared images.
Following~\cite{Nguyen17}, we randomly split the dataset for training and testing sets, where each set contains non-overlapping 206 person identities between the sets. 

\vspace{-9pt}
\paragraph{Evaluation Protocols.}
For SYSU-MM01~\cite{Wu17} benchmark, we follow the evaluation protocol in~\cite{Wu17}. 
We test our model in all-search and indoor-search settings, where the gallery sets for the former include images captured by all four visible cameras, and the latter includes two indoor ones.
For RegDB~\cite{Nguyen17} benchmark, we evaluate our model on infrared to visible and visible to infrared setting, where the former retrieve infrared images from visible ones, and the latter retrieves visible ones from infrared ones.
For both benchmarks, we adopt the cumulative matching characteristics (CMC) and mean average precision (mAP) as evaluation metrics.

\begin{table}
\centering
 \resizebox{0.485\textwidth}{!}{ 
\begin{tabular}{l|cc|cc} 
\hline
\multirow{3}{*}{Methods} & \multicolumn{4}{c}{SYSU-MM01} \\
\cline{2-5}
 & \multicolumn{2}{c|}{Single-shot} & \multicolumn{2}{c}{Multi-shot} \\
\cline{2-5}
                  & Rank-1  &  mAP & Rank-1  &  mAP   \\ 
\hline
\hline
 Base        &  70.58 & 68.24 & 75.58  & 62.91 \\ 
Base+IntraPM      &  72.72 & 69.84 & 76.82  & 64.64 \\ 
Base+InterPM &  75.61  &  71.79  &  78.64  &  67.72  \\ 
Base+IntraPM + InterPM &  75.86  &  72.71  &  79.05  &  68.80  \\ 
\textbf{Ours}   & \textbf{77.78} & \textbf{74.62} & \textbf{80.54} & \textbf{69.84}    \\
\hline
\end{tabular}
}
\vspace{-3pt}
\caption{\textbf{Ablation study for the different components of our method on the SYSU-MM01 dataset~\cite{Wu17}.}
}\label{tab:tab2}
\vspace{-9pt}
\end{table}

\begin{figure}[t]
	\centering
	\includegraphics[width=0.99 \linewidth]{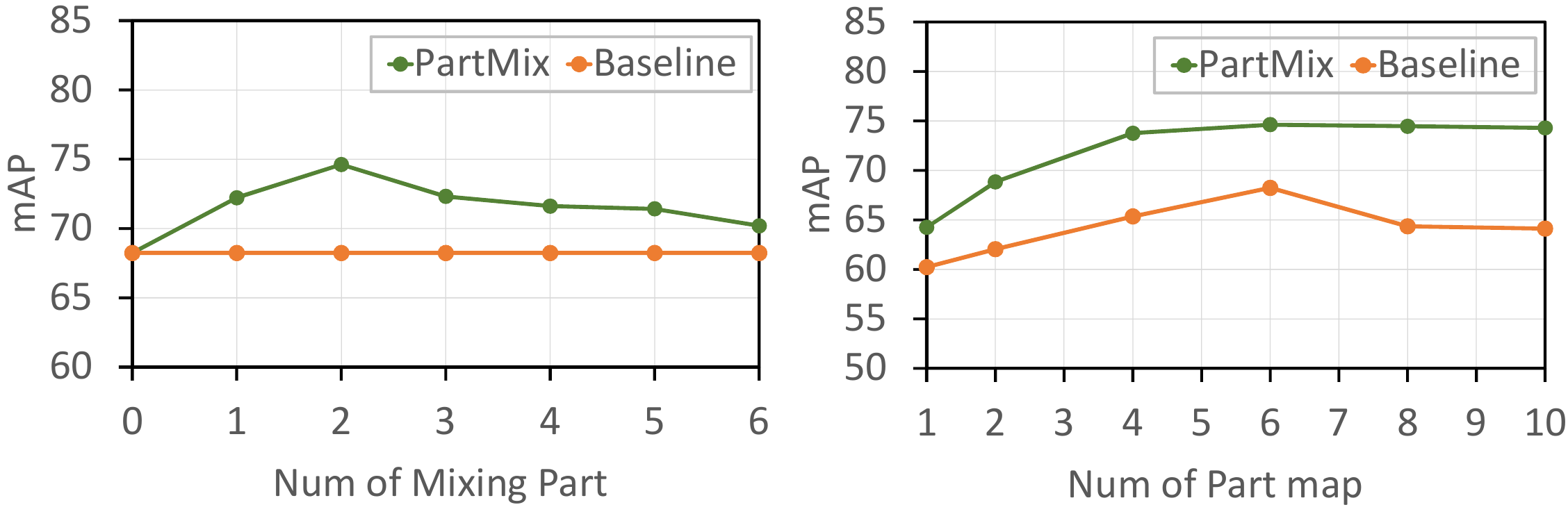}
 \vspace{-3pt}
		\caption{\textbf{Ablation study for part mixing with different numbers of the mixed part (left) and number of part maps (right).}}
	\label{fig:f5}
\vspace{-9pt}
\end{figure}

\subsection{Ablation Study}\label{sec:42}
In this section, we conduct ablation analyses to investigate the effectiveness of our framework.
In this ablation study, all experiments are conducted on SYSU-MM01 dataset~\cite{Wu17}.

\vspace{-9pt}
\paragraph{Effectiveness of Each Component.}
Here we analyze the key components of PartMix, including intra-modality part mixing (IntraPM), inter-modality part mixing (InterPM), and entropy-based mining (EM).
For the fair comparison, we utilize MPANet~\cite{Wu21} as our baseline (Base) for all experiments.
As summarized in \tabref{tab:tab2}, each component helps to boost performance.
Starting from the baseline, adding IntraPM improves the performance which indicates that IntraPM effectively mitigates intra-modality variation (\eg{} human pose variation) thanks to samples synthesized by mixing part information extracted from diverse human poses.
When the InterPM is added, we observe a significant improvement, which confirms that the InterPM effectively synthesizes the unseen combination of human parts across the modalities, and thus mitigates the overfitting to the specific parts and modality, simultaneously.
We also note that EM also brings the performance gain for VI-ReID by mining the reliable samples among initial positive and negative samples.

\vspace{-9pt}
\paragraph{Number of Mixing Parts.}
In our experiment, we mix the part descriptors $B$ times using \equref{equ:4}.
Note that we set the number of part maps as 6 in this experiment.
We analyze the quantitative comparison with a different number of mixed parts $B$ as shown in \figref{fig:f5}. 
The result shows that ours with various numbers of mixed parts, which shows the effectiveness of our PartMix. 
We consider ours with $B = 0$ as the baseline~\cite{Wu21}.
The result with $B = 6$ is that positive and negative samples are set to the counterpart modality samples within the same and different identities, respectively.
The performance has improved as $B$ is increased from 1, and after achieving the best performance at $B = 2$ the rest showed slightly improved performance.
The lower performance with the larger number of mixed parts indicates that easy samples in which the combination of human parts is not significantly different from samples with a different identity in mini-batch can actually be less effective in regularization.
Since the result with $B = 2$ has shown the best performance on the SYSU-MM01 dataset~\cite{Wu17} and RegDB dataset~\cite{Nguyen17}, we set $B$ as 2 for all experiments.

\begin{table}
\centering
 \resizebox{0.46\textwidth}{!}{ 
\begin{tabular}{l|cc|cc} 
\hline
\multirow{3}{*}{Methods} & \multicolumn{4}{c}{SYSU-MM01} \\
\cline{2-5}
 & \multicolumn{2}{c|}{Single-shot} & \multicolumn{2}{c}{Multi-shot} \\
\cline{2-5}
                  & Rank-1  &  mAP & Rank-1  &  mAP   \\ 
\hline
\hline
 Base        &  70.58   & 68.24 & 75.58 &  62.91 \\ 

Base+MixUp~\cite{Zhang18}         & 51.48 & 46.25 & 58.30 & 38.48             \\ 

Base+Manifold~\cite{Verma19}         & 71.25 & 67.74 & 76.72 & 62.39             \\ 
Base+CutMix~\cite{Yun19}         & 73.35  & 70.69 & 77.03 & 64.76            \\ 

\textbf{Ours}  & \textbf{77.78} & \textbf{74.62} & \textbf{80.54} &  \textbf{69.84}   \\
\hline
\end{tabular}
}
\vspace{-3pt}
\caption{\textbf{Ablation study for comparison with other regularization methods on the SYSU-MM01 dataset~\cite{Wu17}.}
}\label{tab:tab3}
\vspace{-8pt}
\end{table}

\vspace{-9pt}
\paragraph{Number of Part Maps.}
In \figref{fig:f5}, we evaluate our model with the different number of part maps. 
In all experiments, we set the number of mixed parts to 1/3 of the number of part maps.
The results show that our PartMix consistently boosts the performance of mAP in single-shot all-search on the SYSU-MM01 dataset. 
For $M = 1$, we exploit training samples with the same and different identities in the mini-batch as positive and negative samples, respectively.
We consider ours with $M = 1$ as contrastive learning with the global descriptor.
It shows that contrastive learning effectively regularizes the model, and thus the model can mitigate modality discrepancy. Specifically, as $M$ is increased from 4, the performance of our model converges to high mAP. 
These results indicate that our PartMix can consistently capture diverse and discriminative human parts with only a small number of part maps.
Since the result with $M=6$ has shown the best performance of mAP in single-shot all-search on the SYSU-MM01 dataset, 
we set $M=6$ for the remaining experiments.

\begin{table*}[t]
    \centering
    \scalebox{0.88}{ 
    \begin{tabular}{c|cc|cc|cc|cc|cc|cc}
\hline  
\multirow{4}{*}{Method}    & \multicolumn{8}{c|}{SYSU-MM01~\cite{Wu17}}    & \multicolumn{4}{c}{\multirow{2}{*}{RegDB~\cite{Nguyen17}}}  \\
\cline{2-9}
    & \multicolumn{4}{c|}{All-Search}       & \multicolumn{4}{c|}{Indoor-Search}         & \multicolumn{4}{c}{}  \\
\cline{2-13}
& \multicolumn{2}{c|}{Single-Shot} & \multicolumn{2}{c|}{Multi-Shot} & \multicolumn{2}{c|}{Single-Shot} & \multicolumn{2}{c|}{Multi-Shot} & \multicolumn{2}{c|}{infrared to visible} & \multicolumn{2}{c}{visible to infrared} \\
\cline{2-13}
    &   Rank-1 & mAP & Rank-1 & mAP   &   Rank-1 & mAP & Rank-1 & mAP & Rank-1 & mAP & Rank-1 & mAP \\ \hline \hline

Two-stream~\cite{Wang19_2}  & 11.65 & 12.85 & 16.33 & 8.03   &   15.60 & 21.49 & 22.49 & 13.92 & - & - & - & - \\ 
One-stream~\cite{Wang19_2} &  12.04 & 13.67 & 16.26 & 8.59   &   16.94 & 22.95 & 22.62 & 15.04 & - & - & - & - \\ 
Zero-Padding~\cite{Wang19_2} & 14.80 & 15.95 & 19.13 & 10.89  &   20.58 & 26.92 & 24.43 & 18.86
& 16.7 & 17.9 & 17.8 & 18.9 \\ 
cmGAN~\cite{Dai18} &  26.97 & 27.80 & 31.49 & 22.27  &   31.63 & 42.19 & 37.00 & 32.76 
& - & - & - & - \\ 
 D\textsuperscript{2}RL~\cite{Wang19} & 28.90 & 29.20 & - & -   &   - & - & - & - 
 & - & - & 43.4 & 44.1 \\ 
JSIA-ReID~\cite{Wang20_3}  & 38.10 & 36.90 & 45.10 & 29.50  &   43.80 & 52.90 & 52.70 & 42.70
& 48.1 & 48.9 & 48.5 & 49.3 \\ 
AlignGAN~\cite{Wang19_2} &  42.40 & 40.70& 51.50 & 33.90  &   45.90 & 54.30 & 57.10 & 45.30 
& 56.3 & 53.4 & 57.9 & 53.6 \\ 
DFE~\cite{Hao19_2}  &48.71 & 48.59 & 54.63 & 42.14  &   52.25 & 59.68 & 59.62 & 50.60 
& 68.0 & 66.7 & 70.2 & 69.2 \\ 
XIV-ReID~\cite{Li20}  & 49.92 & 50.73 &    & - &   - & - & - & - 
& 62.3 & 60.2 & - & - \\ 
CMM+CML~\cite{Ling20} & 51.80 & 51.21 & 56.27 & 43.39  &   54.98 & 63.70 & 60.42 & 53.52 
& 59.8 & 60.9 & - & - \\ 
SIM~\cite{Jia21} & 56.93 & 60.88 & - & -   &   - & - & - & - 
& 75.2 & 78.3 & 74.7 & 75.2 \\ 
CoAL~\cite{Wei20}  & 57.22 & 57.20 & - & -   &   63.86 & 70.84 & - & - 
& 74.1 & 69.9 & - & - \\ 
DG-VAE~\cite{Pu20}  & 59.49 & 58.46 & - & -   &  -& - & - & - 
& - & - & 73.0 & 71.8 \\ 
cm-SSFT~\cite{Lu20} & 61.60 & 63.20 & 63.40 & 62.00  &   70.50 & 72.60 & 73.00 & 72.40 
& 71.0 & 71.7 & 72.3 & 72.9 \\ 
SMCL~\cite{Wei21}   & 67.39 & 61.78 &72.15 & 54.93   &  68.84  & 75.56 & 79.57 & 66.57 
& 83.05& 78.57& 83.93& 79.83\\ 
MPANet~\cite{Wu21}   & 70.58 & 68.24 &75.58 & 62.91   &  76.74  & 80.95 & 84.22 & 75.11 
& 82.8 & 80.7 & 83.7 & 80.9 \\ 
MSCLNet~\cite{Zhang22_2}  & 76.99  & 71.64  & - & - & 78.49 & 81.17 & &
& 83.86 & 78.31 & 84.17 & 80.99 \\
\hline
\bf{Ours} &   \textbf{77.78} & \textbf{74.62} & \textbf{80.54} &  \textbf{69.84}   &  \textbf{81.52} &  \textbf{84.38} & \textbf{87.99} &  \textbf{79.95}
& \textbf{84.93} & \textbf{82.52} & \textbf{85.66} & \textbf{82.27}\\ 
\hline 
\end{tabular}
}
\vspace{-3pt}
\caption{\textbf{Quantitative evaluation on SYSU-MM01 dataset~\cite{Wu17} and RegDB dataset~\cite{Nguyen17}.}
For evaluation, we measure Rank-1 accuracy(\%) and mAP(\%). 
Our results show the best results in terms of Rank-1 accuracy and mAP.
\vspace{-9pt}
}\label{tab:1} 
\end{table*}

\vspace{-3pt}
\subsection{Comparison to Other Regularization Methods}
In this section, we validate the effectiveness of our PartMix through the comprehensive comparison with other regularization methods, including MixUp~\cite{Zhang18}, Manifold MixUp~\cite{Verma19}, and CutMix~\cite{Yun19}.
\tabref{tab:tab3} shows PartMix significantly outperforms all other regularization methods~\cite{Zhang18, Verma19, Yun19}. 
Interestingly, the MixUp method~\cite{Zhang18} highly degrades the performance for part-based VI-ReID.
It demonstrates that simply applying MixUp to the part-based VI-ReID degrades the localization ability of the model due to ambiguous and unnatural patterns in mixed images, and thus 
the model fails to distinguish the different person identities as done in the literature~\cite{Yun19, Venkataramanan22}.
On the other hand, Manifold MixUp~\cite{Verma19} shows slightly improved rank-1 accuracy, but achieves lower performance in mAP than the baseline.
The result shows that it still inherits the limitation of the global mixture model~\cite{Zhang18, Verma19, Shen22}.
Although CutMix~\cite{Yun19} achieves improved performance than the baseline~\cite{Wu21}, it achieves relatively lower performance than ours by 3.93\% and 5.08\% mAP in single-shot all-search and multi-shot all-search on the SYSU-MM01 dataset.
It demonstrated that our method effectively alleviates the overfitting to the specific part and modality in part-based VI-ReID.
Based on all these evaluation and comparison results, we can confirm the effectiveness of our methods.

\subsection{Comparison to Other Methods}
In this section, we evaluate our framework through comparison to state-of-the-art methods for VI-ReID, including Two-stream~\cite{Wang19_2}, One-stream~\cite{Wang19_2}, Zero-Padding~\cite{Wang19_2}, cmGAN~\cite{Dai18}, D\textsuperscript{2}RL~\cite{Wang19}, JSIA-ReID~\cite{Wang20_3}, 
AlignGAN~\cite{Wang19_2}, DFE~\cite{Hao19_2}, XIV-ReID~\cite{Li20}, CMM+CML~\cite{Ling20}, SIM~\cite{Jia21}, CoAL~\cite{Wei20}, DG-VAE~\cite{Pu20}, cm-SSFT~\cite{Lu20}, SMCL~\cite{Wei21},  MPANet~\cite{Wu21}, and MSCLNet~\cite{Zhang22_2}. 

\vspace{-9pt}
\paragraph{Results on SYSU-MM01 dataset.}
We evaluate our PartMix on SYSU-MM01 benchmark~\cite{Wu17} as provided in \tabref{tab:1}.
PartMix achieves the Rank-1 accuracy of $77.78\%$ and mAP of $74.62\%$
in all-search with single-shot mode, improving the Rank-1 accuracy by $0.79\%$ and mAP by $2.98\%$ over the MSCLNet~\cite{Zhang22_2}.
In indoor-search with single-shot mode, our PartMix outperforms the MSCLNet~\cite{Zhang22_2} by Rank-1 accuracy of $3.03\%$ and mAP of $3.21\%$.

\vspace{-9pt}
\paragraph{Results on RegDB dataset.}
We also evaluate our method on RegDB benchmark~\cite{Nguyen17}.
As shown in~\tabref{tab:1}, PartMix records state-of-the-art results with the Rank-1 accuracy of $84.93\%$ and mAP of $82.52\%$ in infrared to visible and the Rank-1 accuracy of $85.66\%$ and mAP of $82.27\%$ in visible to infrared mode.
Our PartMix outperforms the Rank-1 accuracy by $1.07\%$ and mAP by $4.21\%$ 
in infrared to visible mode and the Rank-1 accuracy by $1.49\%$ and mAP by $1.28\%$ in visible to infrared mode over the MSCLNet~\cite{Zhang22_2}.

\vspace{-3pt}
\section{Conclusion}
\label{sec:conclu}
In this paper, we have presented a novel data augmentation technique, called PartMix, that generates part-aware augmented samples by mixing the part descriptors. 
We introduce a novel sample generation method to synthesize the positive and negative samples and an entropy-based mining strategy to select reliable positive and negative samples to regularize the model through the contrastive objective.
We have shown that PartMix achieves state-of-the-art performance over the existing methods on several benchmarks.

\vspace{-9pt}
\paragraph{Acknowledgements.}
This research was supported by the Yonsei Signature Research Cluster Program of 2022 (2022-22-0002) and the KIST Institutional Program (Project No.2E32283-23-064).

{\small
\bibliographystyle{ieee_fullname}
\bibliography{egbib}
}

\clearpage
\newpage
\appendix
\setcounter{table}{0}
\renewcommand{\thetable}{A\arabic{table}}
\setcounter{figure}{0}
\renewcommand{\thefigure}{A\arabic{figure}}
\section*{Appendix}

\input{supple_camera_ready}

\clearpage

\end{document}

%% file: supple_camera_ready.tex
\def\eg{\emph{e.g}.,}
\def\ie{\emph{i.e}.}
\def\etal{\emph{et al.}}

\crefname{section}{Sec.}{Secs.}
\Crefname{section}{Section}{Sections}
\Crefname{table}{Table}{Tables}
\crefname{table}{Tab.}{Tabs.}

\def\cvprPaperID{10469} 
\def\confName{CVPR}
\def\confYear{2023}

In this supplementary material, we provide additional experimental results, implementation details, and qualitative results to complement the main paper.

\section{t-SNE Visualization}
\begin{figure*}[t]
	\centering
	\includegraphics[width=0.99 \linewidth]{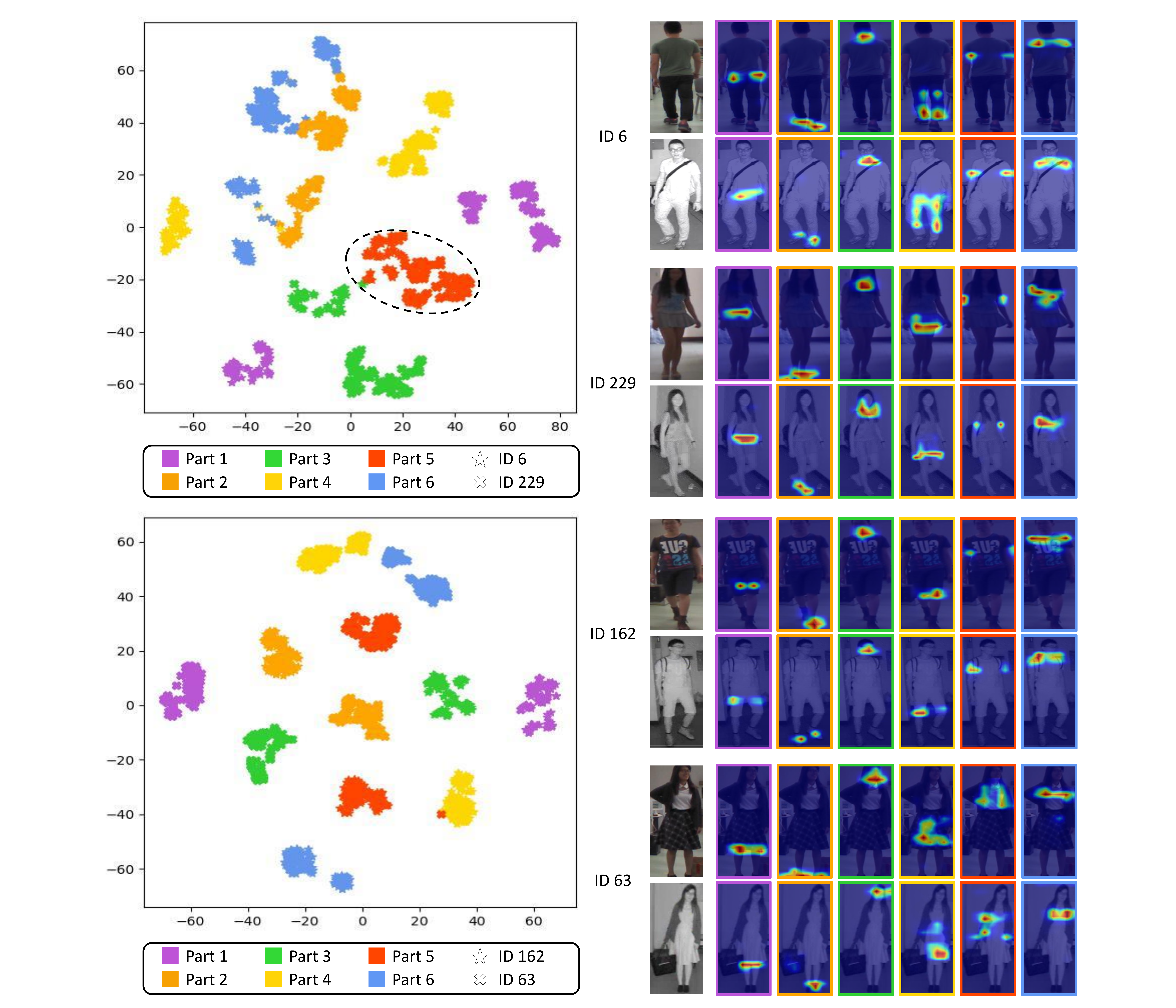}
		\caption{\textbf{Visualization on the feature distribution of part descriptor with different identity images.} Data projection in 2-D space is attained by t-SNE based on the feature representation. Each color represents the different human parts.
		Our PartMix effectively clusters the same human part information (\eg{} short sleeve) in the same group (represented using a dotted circle), while the different human parts are divided into different groups.
		}
	\label{fig:sub_f2}
\end{figure*}

\begin{figure*}[t]
	\centering
	\includegraphics[width=0.99 \linewidth]{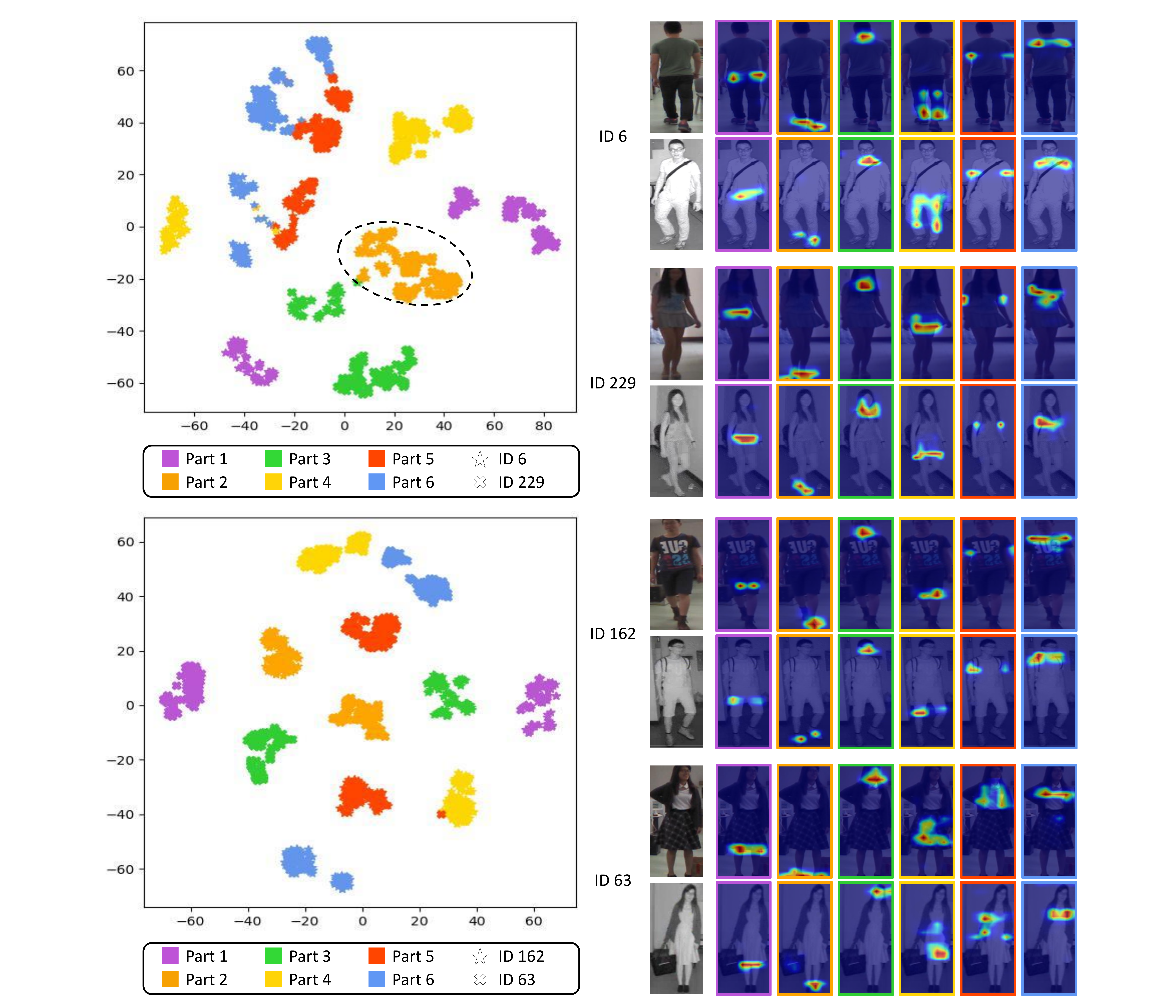}
		\caption{\textbf{Visualization of the feature distribution of part descriptor with different identity images.} The details are the same as above.}
	\label{fig:sub_f3}
\end{figure*}

\paragraph{Visualization for different identities images.}
To explain the effectiveness of our PartMix, we show the feature distribution of part descriptor with different identities in
\figref{fig:sub_f2}.
For visualizing the feature distribution, the complex feature distributions are transformed into two-dimensional points based on t-SNE~\cite{Van08}.
Each color represents the $M$ different part maps. 
We can confirm that t-SNE visualization of part descriptors that have different semantic meanings are clustered into distinct groups.
And we can also find that the part descriptor with the same human part information (\eg{} short sleeve) are clustered into the same groups.
In \figref{fig:sub_f3}, we visualize an additional example for the feature distribution of part descriptors with different identities.
These two images do not share the human parts information, and thus our PartMix effectively divides part descriptors into different groups. 
By this visualization, we can demonstrate that our PartMix can capture different human part information and synthesize unseen combination of human parts (\ie{} the unseen identity), improving generalization ability on unseen identity as demonstrated in the Sec 4.4 of the main paper. 
In addition, it can distinguish the different person identities through the combination of human parts.

\section{Loss functions}
Following the baseline~\cite{Wu21}, we adopt several losses, including modality learning loss $\mathcal{L}_{\mathrm{ML}}$, 
modality specific ID loss $\mathcal{L}_{\mathrm{sid}}$, center cluster loss $\mathcal{L}_{\mathrm{cc}}$, 
and identity classification loss $\mathcal{L}_{\mathrm{id}}$.
In this section, we describe these losses in detail.

\vspace{-9pt}
\paragraph{Modality Learning Loss.}
Modality learning loss~\cite{Wu21} aims to encourage the modality-specific classifier to estimate consistent classification scores for the same identity features regardless of the modality.
We make the classification scores of visible (infrared) person descriptors estimated by the visible (infrared) and mean infrared (visible) specific classifier to be similar through the KL divergence, and thus the model learns modality invariant person descriptors.

\begin{equation}\label{equ:10}
\begin{split}
&\mathcal{L}_\mathrm{ML} = \sum^{N_{v}}_{w=1} d_{KL} ( \mathcal{C}_{v}(d^{v}_{w})||\tilde{\mathcal{C}}_{r}(d^{v}_{w})) \\
&\;\;\;\;\;\;\;\;\;\;\;\;\;\;\;\;\;\;\;\;\;\;\;\;\;\;\;+ \sum^{N_{r}}_{q=1} d_{KL} ( \mathcal{C}_{r}(d^{r}_{q})||\tilde{\mathcal{C}}_{v}(d^{r}_{q})),
\end{split}
\end{equation}
where  $\mathcal{C}_{v}(\cdot)$, $\mathcal{C}_{r}(\cdot)$ denote visible and infrared classifiers, and the mean classifiers of those ones are $\tilde{\mathcal{C}}_{v}(\cdot)$, $\tilde{\mathcal{C}}_{r}(\cdot)$, respectively.

\vspace{-9pt}
\paragraph{Modality Specific ID Loss.}
For modality learning loss, we train the modality-specific classifiers to learn modality-specific knowledge from visible and infrared person descriptors such that
\begin{equation}\label{equ:11}
\begin{split}
&\mathcal{L}_\mathrm{sid} = - \frac{1}{N_{v}} \sum^{N_{v}}_{w=1} y_{w}^{v} \log (\mathcal{C}_{v}(d^{v}_{w})) \\
&\;\;\;\;\;\;\;\;\;\;\;\;\;\;\;\;\;\;\;\;\;\;\;\;\;\;\;\;\;\;\;\;\; - \frac{1}{N_{r}} \sum^{N_{r}}_{q=1} y^{r}_{q} \log(\mathcal{C}_{r}(d^{r}_{q})),
\end{split}
\end{equation}
where $\mathcal{C}_{v}$ and $\mathcal{C}_{r}$ are visible and infrared classifier.

\vspace{-9pt}
\paragraph{Center Cluster Loss.}
To enhance the discriminative power of the person descriptor, 
we adopt center cluster loss~\cite{Wu21} to penalize the distances between the person descriptors and their corresponding identity centers.
\begin{equation}\label{equ:12}
\begin{split}
&\mathcal{L}_\mathrm{cc} = \frac{1}{N} \sum^{N}_{i=1} ||f^{t}_{i} - z_{y_{i}}||_{2} \\
&\;\;\;\;\;\;\;\; + \frac{2}{P(P-1)}\sum^{P-1}_{k=1}\sum^{P}_{d=k+1}[\rho - ||z_{y_{k}}-z_{y_{d}}||_{2}]_{+},
\end{split}
\end{equation}
where $z_{y_{i}}$,$z_{y_{k}}$, and $z_{y_{d}}$ is the mean descriptor that correspond to the $y_{i}$, $y_{k}$, and $y_{d}$ identity in mini-batch, $P$ is the number of identity in the mini-batch, and $\rho$ is the least margin between the centers.

\vspace{-9pt}
\paragraph{ID Loss}
To learn identity-specific feature representation across the modalities, we adopt cross-entropy loss between the identity probabilities and their ground-truth identities as follows:
\begin{equation}\label{equ:3}
  \mathcal{L}_\mathrm{id} = -\frac{1}{N^{v}} \; \sum_{i=1}^{N^{v}} y^{v}_{i} \log(\mathcal{C}(d^{v}_{i}))-\frac{1}{N^{r}} \; \sum_{i=1}^{N^{r}} y^{r}_{i} \log(\mathcal{C}(d^{r}_{i})),
\end{equation}
where $\mathcal{C}(\cdot)$ is an identity classifier.

\section{Implementation Details}

\paragraph{Training Details.}
To train our network, we first conduct warm up the baseline~\cite{Wu21} for 20 epochs, to stabilize the part detector at the early stage of training and boost the convergence of training. 
For a fair comparison with the baseline~\cite{Wu21}, we then optimize the model for 100 epochs using overall losses.
We also adopt random cropping, random horizontal flipping, and random erasing~\cite{Zhong20_2} for data augmentation.
We set 128 images for each mini-batch.
For each mini-batch, we randomly sample 8 images with 16 identities and the images are re-sized as 384$\times$128. 
We select positive samples and negative samples through the entropy-based mining module.
For each training sample, we set the number of positive ${U}'$ and negative samples ${Q}'$ as 2 and 20.
To optimize the model, we utilize the Adam optimizer, where the initial learning rate is set to $3.5 \times {10}^{-4}$, which decays at 80$^{th}$ and $120^{th}$ epoch with a decay factor of 0.1.
Through the cross-validation using grid-search, we set the hyper-parameters $\lambda_{\mathrm{sid}}$, $\lambda_{\mathrm{ML}}$, $\lambda_{cont}$, and $\lambda_{aid}$ as 0.5, 2.5, 0.5, and 0.5, respectively.
The proposed method was implemented in the Pytorch library~\cite{Paszke17}.
We conduct all experiments using a single RTX A6000 GPU.

\section{Other Regularization Methods Details}

\paragraph{Mixup~\cite{Zhang18}.}
Following the work~\cite{Zhang18}, we synthesize the mixed images by linearly interpolating image and label pairs such that 
\begin{equation}\label{equ:13}
\begin{split}
&\tilde{x} =  \lambda x^{1} + (1-\lambda) x^{2},\\
&\tilde{y} = \lambda y^{1} + (1- \lambda ) y^{2},
\end{split}
\end{equation}
where $x^{1}, x^{2}$ are randomly sampled images in mini-batch regardless of their modality, $y^{1}, y^{2}$ are its corresponding identity, and $\lambda$ is the combination ratio sampled from the beta distribution $Beta(\alpha, \alpha)$, where the $\alpha$ is set to 1.

\vspace{-9pt}
\paragraph{Manifold MixUp~\cite{Verma19}.}
We also synthesize the mixed training samples using Manifold MixUp~\cite{Verma19} that applies MixUp~\cite{Zhang18} in the hidden feature space as follows : 
\begin{equation}\label{equ:14}
\begin{split}
&\tilde{x} =  \lambda \mathcal{E}_{g}(x^{1}) + (1-\lambda) \mathcal{E}_{g}(x^{2}),\\
&\tilde{y} = \lambda y^{1} + (1- \lambda ) y^{2},
\end{split}
\end{equation}
 where $\mathcal{E}_{g}(x)$ denotes a forward pass until randomly chosen layer $g$. 
 We also sample the combination ratio $\lambda$ from the beta distribution $\beta(\alpha, \alpha)$, where the $\alpha$ is set as 1.

\vspace{-9pt}
\paragraph{CutMix~\cite{Yun19}.}
We generate training samples with CutMix operation as follows:
\begin{equation}\label{equ:15}
\begin{split}
&\tilde{x} = \mathbf{M} \odot x^{1} + (\mathbf{1}-\mathbf{M}) \odot x^{2},\\
&\tilde{y} = \lambda y^{1} + (1- \lambda ) y^{2},
\end{split}
\end{equation}
where $\mathbf{M}$ is a binary mask, $\mathbf{1}$ is a binary mask filled with ones, $\odot$ is element-wise multiplication, and the setting of $\lambda$ is identical to Mixup~\cite{Zhang18}.
To sample the mask $\mathbf{M}$, we uniformly sample the bounding box coordinates $\mathbf{B} = (b_{x}, b_{y}, b_{w}, b_{h})$ such that
\begin{equation}\label{equ:16}
\begin{split}
&b_{x} \sim Unif (0, W), b_{w} = W\sqrt{1-\lambda},\\
&b_{y} \sim Unif (0, H), b_{h} = H\sqrt{1-\lambda},
\end{split}
\end{equation}
where $W, H$ is width and height of the person image and $Unif(\cdot, \cdot)$ denotes a uniform distribution.

